\pdfoutput=1

\documentclass{article}

\usepackage[nonatbib, preprint]{nips_2018}

\usepackage[utf8]{inputenc} 
\usepackage[T1]{fontenc}    
\usepackage{hyperref}       
\usepackage{url}            
\usepackage{booktabs}       
\usepackage{amsfonts}       
\usepackage{nicefrac}       
\usepackage{microtype}      
\usepackage[ruled,vlined]{algorithm2e}
\usepackage{mathtools}
\usepackage{gensymb}
\usepackage{graphicx}
\usepackage{subcaption}
\usepackage{float}
\usepackage{epstopdf} 
\usepackage{wrapfig}

\newcommand{\ProjectName}{DeepCube}
\newcommand{\AlgoName}{Autodidactic Iteration}
\newcommand{\AlgoShort}{ADI}

\DeclareMathOperator*{\argmax}{argmax}

\title{Solving the Rubik's Cube Without Human Knowledge}

\usepackage{hyperref}

\makeatletter
\newcommand{\printfnsymbol}[1]{%
  \textsuperscript{\@fnsymbol{#1}}%
}
\makeatother

\author{
  Stephen McAleer\thanks{Equal contribution} \\
  Department of Statistics\\
  University of California, Irvine\\
  \texttt{smcaleer@uci.edu} \\
  \And
  Forest Agostinelli\printfnsymbol{1} \\
  Department of Computer Science \\
  University of California, Irvine\\
  \texttt{fagostin@uci.edu}\\
  \AND
  Alexander Shmakov\printfnsymbol{1} \\
  Department of Computer Science \\
  University of California, Irvine \\
  \texttt{ashmakov@uci.edu} \\
  \And
  Pierre Baldi \\
  Department of Computer Science \\
  University of California, Irvine \\
  \texttt{pfbaldi@ics.uci.edu} \\
}

\begin{document}

\maketitle

\begin{abstract}
A generally intelligent agent must be able to teach itself how to solve problems in complex domains with minimal human supervision. Recently, deep reinforcement learning algorithms combined with self-play have achieved superhuman proficiency in Go, Chess, and Shogi without human data or domain knowledge. In these environments, a reward is always received at the end of the game; however, for many combinatorial optimization environments, rewards are sparse and episodes are not guaranteed to terminate. We introduce \AlgoName{}: a novel reinforcement learning algorithm that is able to teach itself how to solve the Rubik’s Cube with no human assistance. Our algorithm is able to solve 100\% of randomly scrambled cubes while achieving a median solve length of 30 moves --- less than or equal to solvers that employ human domain knowledge. 
\end{abstract}




\section{Introduction}
Reinforcement learning seeks to create intelligent agents that adapt to an environment by analyzing their own experiences. 
Reinforcement learning agents have achieved superhuman capability in a number of competitive games \cite{tesauro1995temporal,mnih2015human,silver2017mastering1,silver2017mastering2}. 
This recent success of reinforcement learning has been a product of the combination of classic reinforcement learning \cite{bellman1957dynamic,sutton_1988,
watkins1992q,sutton1998reinforcement}, deep learning \cite{lecun2015deep,goodfellow2016deep}, and Monte Carlo Tree Search (MCTS) \cite{coulom2006efficient,kocsis2006bandit, herik_2007}. The fusion of reinforcement learning and deep learning is known as deep reinforcement learning (DRL). Though DRL has been successful in many different domains, it relies heavily on the condition that an informatory reward can be obtained from an initially random policy. 
Current DRL algorithms struggle in environments with a high number of states and a small number of reward states such as \textit{Sokoban} and \textit{Montezuma's Revenge}. Other environments, such as short answer exam problems, prime number factorization, and combination puzzles like the Rubik's Cube have a large state space and only a single reward state.

The 3x3x3 Rubik's cube is a classic 3-Dimensional combination puzzle. There are 6 faces, or 3x3x1 planes, which can be rotated $90^{\circ}$ in either direction. The goal state is reached when all stickers on each face of the cube are the same color, as shown in Figures \ref{fig:solvedTop} and \ref{fig:solvedBottom}. The Rubik's cube has a large state space, with approximately $4.3 \times 10^{19}$ different possible configurations. However, out of all of these configurations, only one state has a reward signal: the goal state. Therefore, starting from random states and applying DRL algorithms, such as asynchronous advantage actor-critic (A3C) \cite{mnih2016asynchronous}, could theoretically result in the agent never solving the cube and never receiving a learning signal.

To overcome a sparse reward in a model-based environment with a large state space, we introduce \AlgoName{} (\AlgoShort): an algorithm inspired by policy iteration \cite{bellman1957dynamic,sutton1998reinforcement} for training a joint value and policy network. \AlgoShort{} trains the value function through an iterative supervised learning process. In each iteration, the inputs to the neural network are created by starting from the goal state and randomly taking actions. The targets seek to estimate the optimal value function by performing a breadth-first search from each input state and using the current network to estimate the value of each of the leaves in the tree. Updated value estimates for the root nodes are obtained by recursively backing up the values for each node using a max operator. The policy network is similarly trained by constructing targets from the move that maximizes the value.
After the network is trained, it is combined with MCTS to efficiently solve the Rubik's Cube. We call the resulting solver \ProjectName{}.

\begin{figure}[t]
	\centering
	\includegraphics[width=0.99\textwidth]{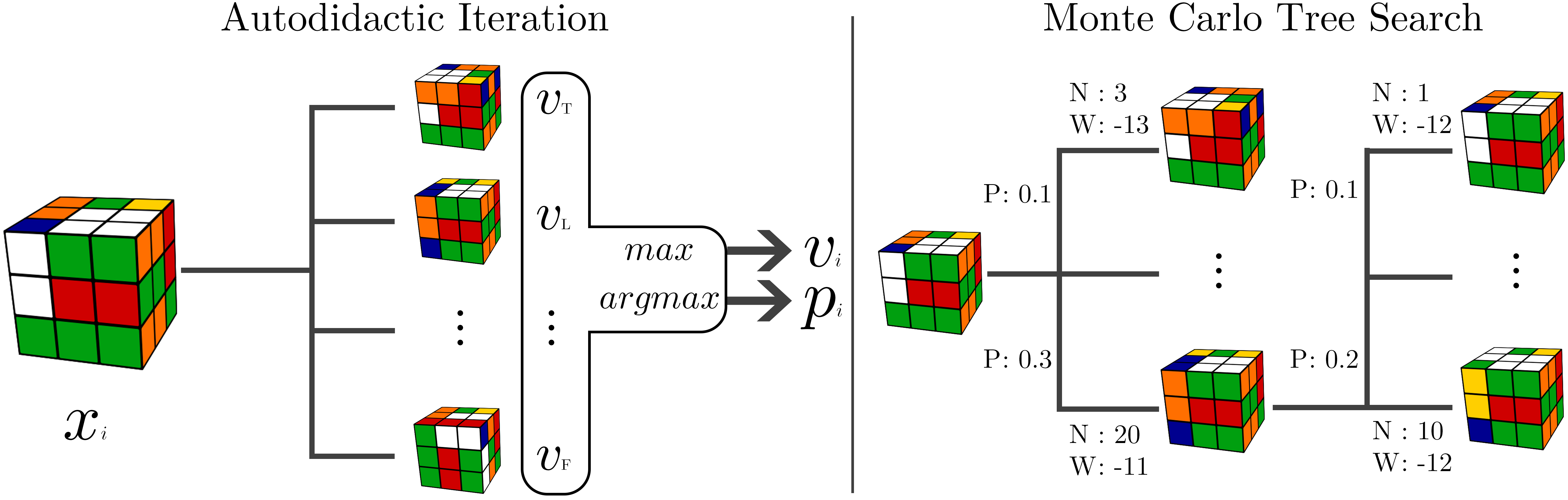}
    \caption{An illustration of \ProjectName{}. The training and solving process is split up into \AlgoShort{} and MCTS. First, we iteratively train a DNN by estimating the true value of the input states using breadth-first search. Then, using the DNN to guide exploration, we solve cubes using Monte Carlo Tree Search. See methods section for more details.}
	\label{fig:overview}
\end{figure}

\section{Related work}
Erno Rubik created the Rubik’s Cube in 1974. After a month of effort, he came up with the first algorithm to solve the cube. Since then, the Rubik's Cube has gained worldwide popularity and many human-oriented algorithms for solving it have been discovered \cite{ruwix}. These algorithms are simple to memorize and teach humans how to solve the cube in a structured, step-by-step manner.

Human-oriented algorithms to solve the Rubik's Cube, while easy to memorize, find long suboptimal solutions. Since 1981, there has been theoretical work on finding the upper bound for the number of moves necessary to solve any valid cube configuration \cite{thistlethwaite_1981,reid_1995,radu_2007,kunkle_2007,rokicki2010twenty}. Finally, in 2014, it was shown that any valid cube can be optimally solved with at most 26 moves in the quarter-turn metric, or 20 moves in the half-turn metric \cite{rokicki2014diameter, rokicki_2014}. The quarter-turn metric treats 180 degree rotations as two moves, whereas the half-turn metric treats 180 degree rotations as one move. For the remainder of this paper we will be using the quarter-turn metric. This upper bound on the number of moves required to solve the Rubik's cube is colloquially known as God's Number.

Algorithms used by machines to solve the Rubik's Cube rely on hand-engineered features and group theory to systematically find solutions. One popular solver for the Rubik's Cube is the Kociemba two-stage solver \cite{kociemba_2018}. This algorithm uses the Rubik's Cube's group properties to first maneuver the cube to a smaller sub-group, after which finding the solution is trivial. Heuristic based search algorithms have also been employed to find optimal solutions. Korf first used a variation of the A* heuristic search, along with a pattern database heuristic, to find the shortest possible solutions \cite{korf_1997}. More recently, there has been an attempt to train a DNN -- using supervised learning with hand-engineered features -- to act as an alternative heuristic \cite{Brunetto_2017}. These search algorithms, however, take an extraordinarily long time to run and usually fail to find a solution to randomly scrambled cubes within reasonable time constraints. Besides hand-crafted algorithms, attempts have been made to solve the Rubik's Cube through evolutionary algorithms \cite{smith2016discovering,lichodzijewski2011rubik}. However, these learned solvers can only reliably solve cubes that are up to 5 moves away from the solution. 


We solve the Rubik's Cube using pure reinforcement learning without human knowledge. In order to solve the Rubik's Cube using reinforcement learning, the algorithm will learn a \textit{policy}. The policy determines which move to take in any given state. Much work has already been done on finding optimal policy functions in classic reinforcement learning \cite{sutton_1988,watkins1992q,williams1992simple,moriarty1999evolutionary}. \AlgoName{} can be seen as a type of policy iteration algorithm \cite{sutton1998reinforcement}. Since our implementation of \AlgoShort{} uses a depth-1 greedy policy when training the network, it can also be thought of as value iteration. Policy iteration is an iterative reinforcement learning algorithm which alternates between a policy evaluation step and a policy improvement step. The policy evaluation step estimates the state-value function given the current policy, while the policy improvement step uses the estimated state-value function to improve the policy. Though many policy iteration algorithms require multiple sweeps through the state space for each step, value iteration effectively combines policy iteration and policy improvement into a single sweep. 
Other works have used neural networks to augment MCTS \cite{guo_2014,silver_2016,anthony_2017}. Our approach is most similar to AlphaGo Zero \cite{silver_2017} and Expert Iteration \cite{anthony_2017}, which also do not use human knowledge. 

\section{The Rubik's Cube}
\begin{figure}[t]
    \centering
    \begin{subfigure}[b]{0.48\textwidth}
    \begin{subfigure}{0.47\textwidth}
        \includegraphics[width=\textwidth]{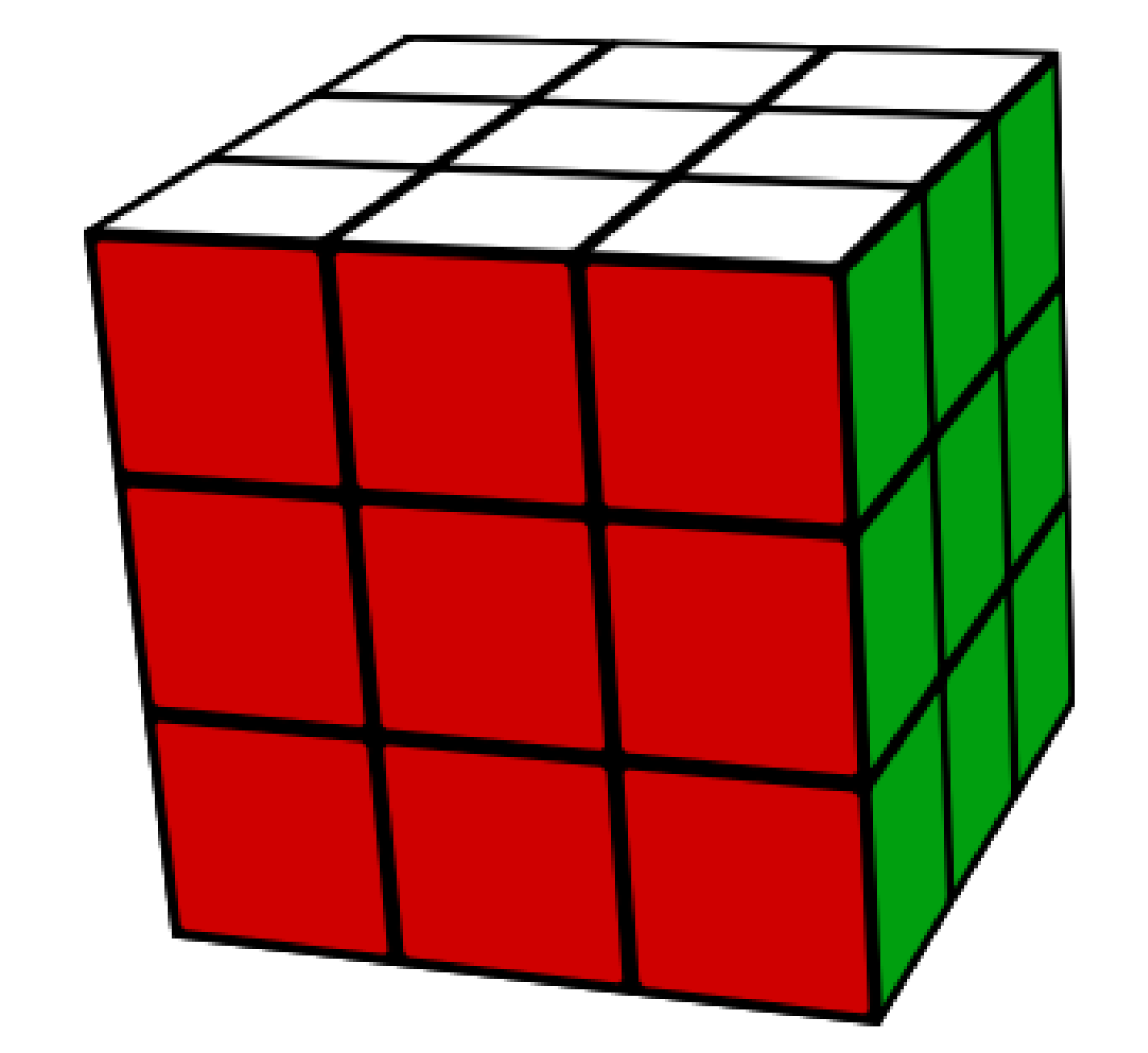}
        \caption{Top view}
        \label{fig:solvedTop}
    \end{subfigure}
    \begin{subfigure}{0.47\textwidth}
        \includegraphics[width=\textwidth]{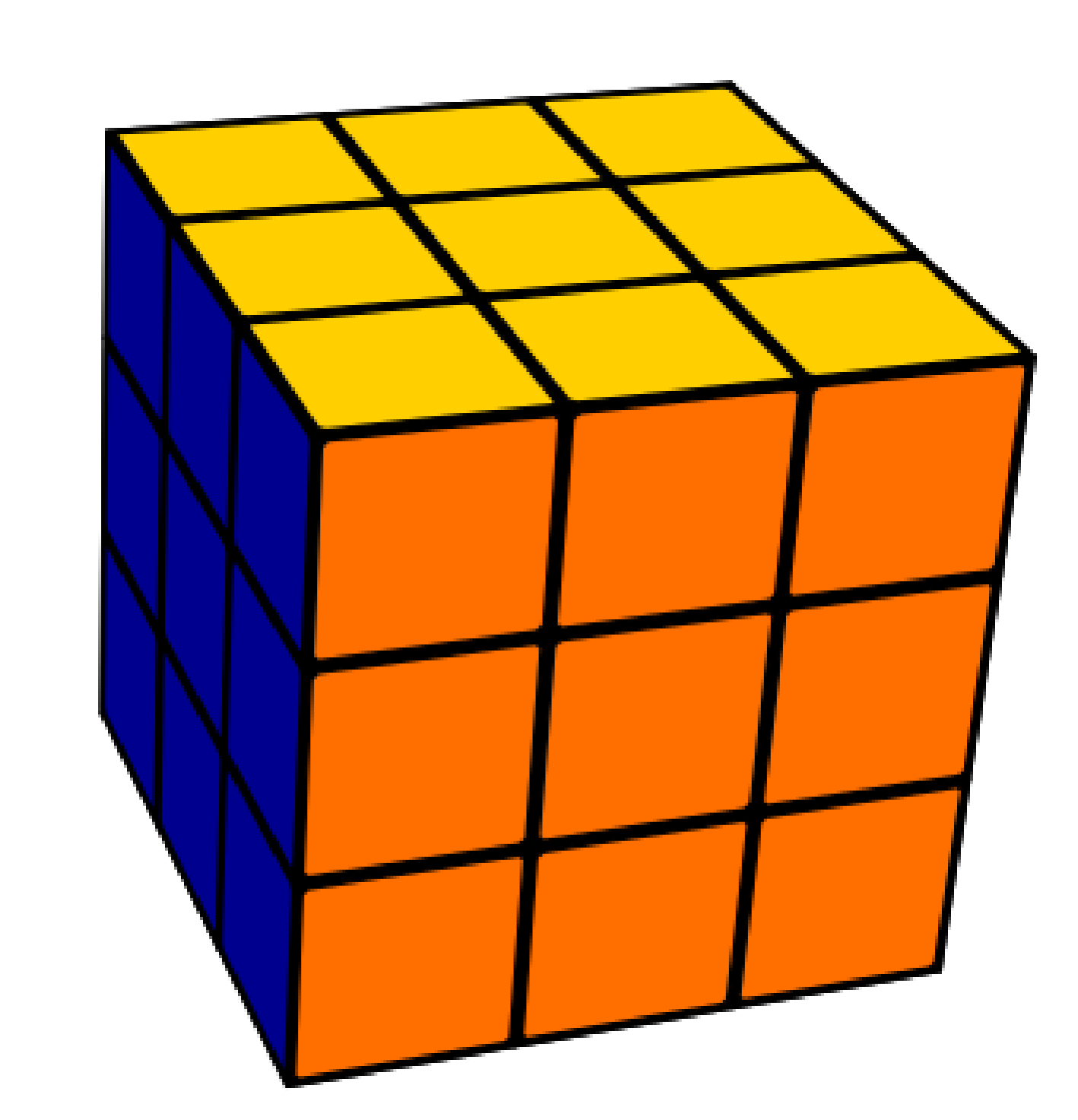}
        \caption{Bottom view}
        \label{fig:solvedBottom}
    \end{subfigure}
    \end{subfigure}
    \begin{subfigure}[b]{0.48\textwidth}
    \begin{subfigure}{0.47\textwidth}
        \includegraphics[width=\textwidth]{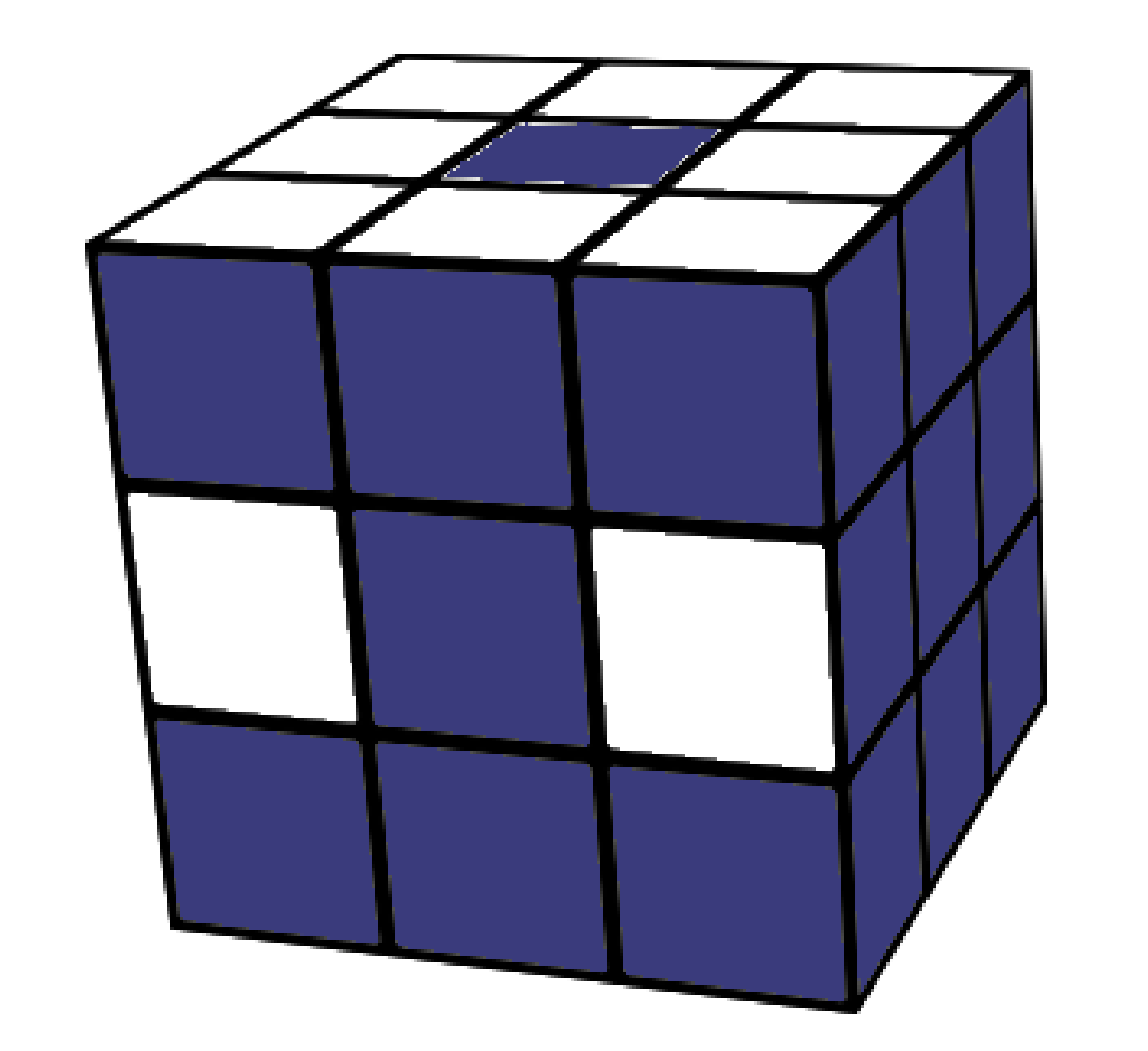}
        \caption{Top view}
        \label{fig:solvedTop_compr}
    \end{subfigure}
    \begin{subfigure}{0.47\textwidth}
        \includegraphics[width=\textwidth]{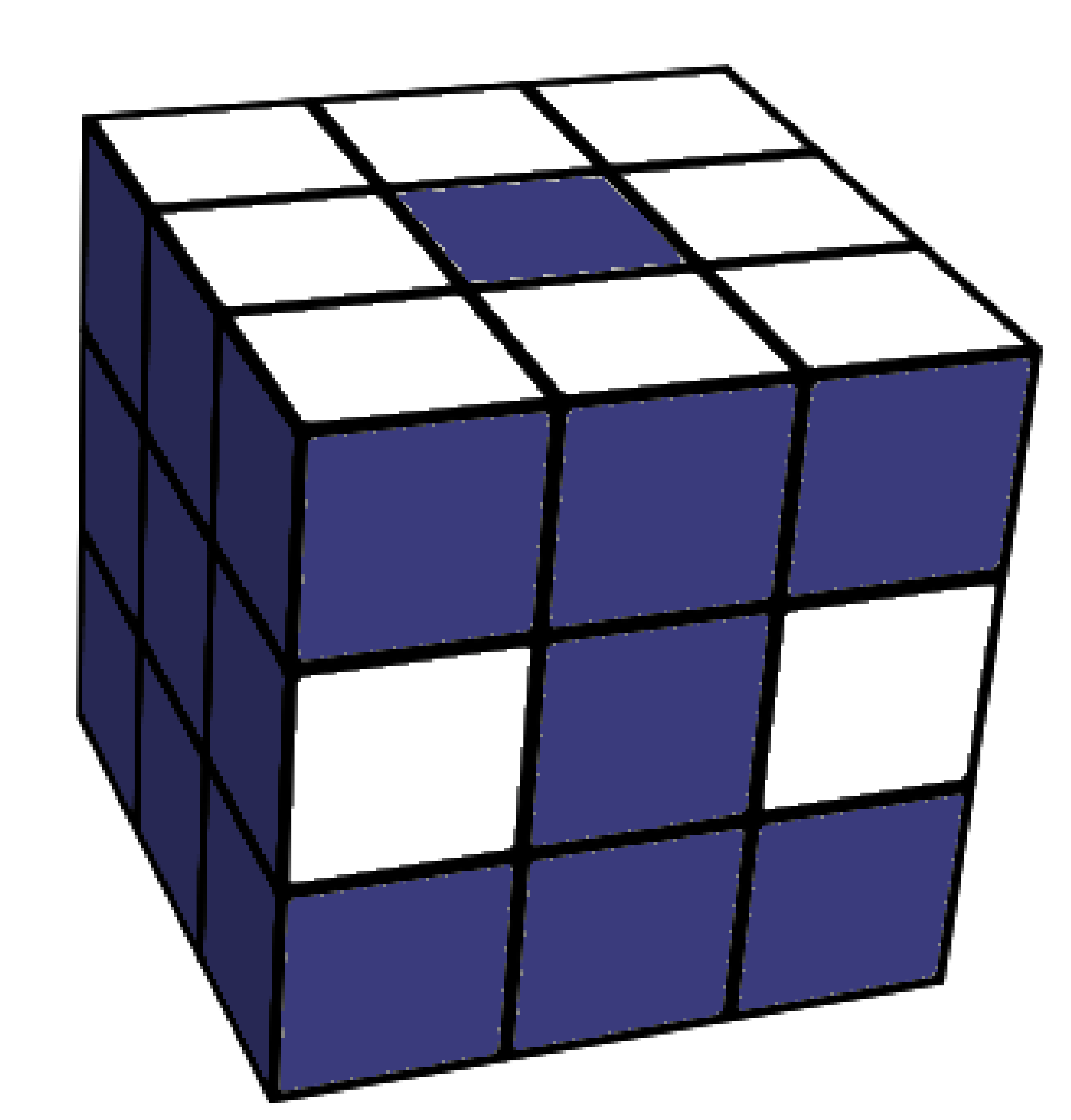}
        \caption{Bottom view}
        \label{fig:solvedBottom_compr}
    \end{subfigure}
    \end{subfigure}
    
    \caption{Visualizations of the Rubik's Cube. (a) and (b) show the solved cube as it appears in the environment. (c) and (d) show the cube reduced in dimensionality for input into the DNN. Stickers that are used by the DNN are white, whereas ignored stickers are dark.}
	\label{fig:solved}
\end{figure}

The Rubik's Cube consists of 26 smaller cubes called cubelets. These are classified by their sticker count: center, edge, and corner cubelets have 1, 2, and 3 stickers attached respectively. There are 54 stickers in total with each sticker uniquely identifiable based on the type of cubelet the sticker is on and the other sticker(s) on the cubelet. Therefore, we may use a one-hot encoding for each of the 54 stickers to represent their location on the cube.

However, because the position of one sticker on a cubelet determines the position of the remaining stickers on that cubelet, we may actually reduce the dimensionality of our representation by focusing on the position of only one sticker per cubelet. We ignore the redundant center cubelets and only store the 24 possible locations for the edge and corner cubelets. This results in a 20x24 state representation which is demonstrated in Figures \ref{fig:solvedTop_compr} and \ref{fig:solvedBottom_compr}. 


Moves are represented using face notation originally developed by David Singmaster: a move is a letter stating which face to rotate. $\textit{F}$, $\textit{B}$, $\textit{L}$, $\textit{R}$, $\textit{U}$, and $\textit{D}$ correspond to turning the \textit{front}, \textit{back}, \textit{left}, \textit{right}, \textit{up}, and \textit{down} faces, respectively. A clockwise rotation is represented with a single letter, whereas a letter followed by an apostrophe represents a counter-clockwise rotation. For example: $\textit{R}$ and $\textit{R'}$ would mean to rotate the \textit{right} face $90^{\circ}$ clockwise and counter-clockwise, respectively.

Formally, the Rubik's Cube environment consists of a set of $4.3 \times 10^{19}$ states $\mathcal{S}$ which contains one special state, $s_{solved}$, representing the goal state. At each timestep, $t$, the agent observes a state $s_t \in \mathcal{S}$ and takes an action $a_t \in \mathcal{A}$ with $\mathcal{A} := \left\lbrace \textit{F, F', \ldots , D, D'} \right\rbrace$. After selecting an action, the agent observes a new state $s_{t+1} = A(s_t, a_t)$ and receives a scalar reward, $R(s_{t+1})$, which is 1 if $s_{t+1}$ is the goal state and $-1$ otherwise.

\section{Methods}
\begin{figure}[t]
	\includegraphics[width=0.99\textwidth]{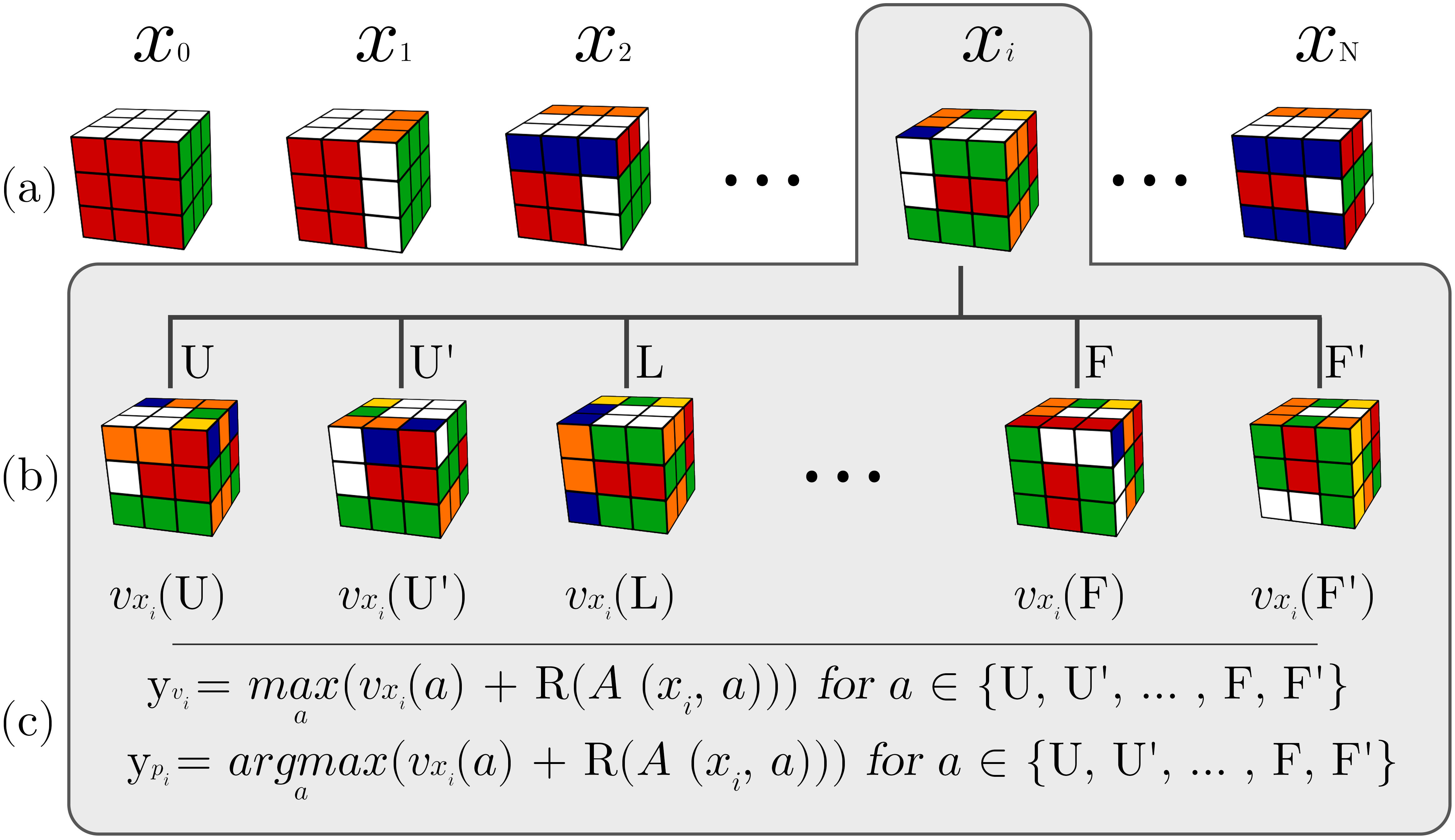}
    \caption{Visualization of training set generation in \AlgoShort{}. (a) Generate a sequence of training inputs starting from the solved state. (b) For each training input, generate its children and evaluate the value network on all of them. (c) Set the value and policy targets based on the maximal estimated value of the children.}
	\label{fig:training}
\end{figure}



We develop a novel algorithm called \AlgoName{} which is used to train a joint value and policy network. Once the network is trained, it is combined with MCTS to solve randomly scrambled cubes. The resulting solver is called \ProjectName{}. 

\subsection{Autodidactic Iteration}
\AlgoShort{} is an iterative supervised learning procedure which trains a deep neural network $f_\theta(s)$ with parameters $\theta$ which takes an input state $s$ and outputs a value and policy pair $(v, \boldsymbol{p})$. The policy output $\boldsymbol{p}$ is a vector containing the move probabilities for each of the $12$ possible moves from that state. Once the network is trained, the policy is used to reduce breadth and the value is used to reduce depth in the MCTS. 
In each iteration of \AlgoShort{}, training samples for $f_\theta$ are generated by starting from the solved cube. This ensures that some training inputs will be close enough to have a positive reward when performing a shallow search. Targets are then created by performing a depth-1 breadth-first search (BFS) from each training sample. The current value network is used to estimate each child's value. The value target for each sample is the maximum value and reward of each of its children, and the policy target is the action which led to this maximal value. Figure \ref{fig:training} displays a visual overview of \AlgoShort{}.

Formally, we generate training samples by starting with $s_{solved}$ and scrambling the cube $k$ times to generate a sequence of $k$ cubes. We do this $l$ times to generate $N = k*l$ training samples $X = [x_{i}]_{i=1}^N$. Each sample in the series has the number of scrambles it took to generate it, $D(x_i)$, associated with it.
Then, for each sample $x_i \in X$, we generate a training target $Y_i = (y_{v_i}, \boldsymbol{y}_{p_i})$. To do this, we perform a depth-1 BFS to get the set of all children states of $x_i$. We then evaluate the current neural network at each of the children states to receive estimates of each child's optimal value and policy: $\forall a \in \mathcal{A}, \ (v_{x_i}(a), \boldsymbol{p}_{x_i}(a)) = f_\theta(A(x_i, a))$. We set the value target to be the maximal value from each of the children $y_{v_i} \gets \max_a(R(A(x_i, a)) + v_{x_i}(a))$, and we set the policy target to be  the move that results in the maximal estimated value $\boldsymbol{y}_{p_i} \gets \operatorname{argmax}_a (R(A(x_i, a)) + v_{x_i}(a))$. We then train $f_\theta$ on these training samples and targets $[x_i, y_{v_i}]_{i=1}^N$ and $[x_i, \boldsymbol{y}_{p_i}]_{i=1}^N$ to receive new neural network parameters $\theta'$. For training, we used the \textit{RMSProp} optimizer \cite{hinton_2014} with a mean squared error loss for the value and softmax cross entropy loss for the policy. Although we use a depth-1 BFS for training, this process may be trivially generalized to perform deeper searches at each $x_i$.

\begin{algorithm}[t]
	\caption{\AlgoName{}}
	\label{training}
	\textbf{Initialization:} $\theta$ initialized using Glorot initialization\\
	\Repeat{$iterations=M$}{
		$X \gets$ N scrambled cubes\\
        \For{$x_i\in X$}{
            \For{$a \in \mathcal{A}$}{
                $(v_{x_i}(a), \boldsymbol{p}_{x_i}(a)) \gets f_\theta(A(x_i, a))$\\
            }
            $y_{v_i} \gets \max_a(R(A(x_i, a)) + v_{x_i}(a))$\\
            $\boldsymbol{y}_{p_i} \gets \operatorname{argmax}_a (R(A(x_i, a)) + v_{x_i}(a))$\\
            $Y_i \gets (y_{v_i}, \boldsymbol{y}_{p_i})$\\
		}
        $\theta' \gets train(f_\theta, X, Y)$ \\
        $\theta \gets \theta'$ \\
	}
\end{algorithm}

\subsubsection*{Weighted samples}
During testing, we found that the learning algorithm sometimes either converged to a degenerate solution or diverged completely. To counteract this, we assigned a higher training weight to samples that are closer to the solved cube compared to samples that are further away from solution. We assign a loss weight to each sample $x_i$, $W(x_i) = \frac{1}{D(x_i)}$. We didn't see divergent behavior after this addition.

\subsection{Solver}
We employ an asynchronous Monte Carlo Tree Search augmented with our trained neural network $f_\theta$ to solve the cube from a given starting state $s_0$. We build a search tree iteratively by beginning with a tree consisting only of our starting state, $T = \left\lbrace s_0 \right\rbrace$. We then perform simulated traversals until reaching a leaf node of $T$. Each state, $s \in T$, has a memory attached to it storing: $N_s(a)$, the number of times an action $a$ has been taken from state $s$, $W_s(a)$, the maximal value of action $a$ from state $s$, $L_s(a)$, the current virtual loss for action $a$ from state $s$, and $P_s(a)$, the prior probability of action $a$ from state $s$. 

Every simulation starts from the root node and iteratively selects actions by following a tree policy until an unexpanded leaf node, $s_\tau$, is reached. The tree policy proceeds as follows: for each timestep $t$, an action is selected by choosing, $A_t = \argmax_{a} U_{s_t}(a) + Q_{s_t}(a)$ where $U_{s_t}(a) = c P_{s_t}(a) {\sqrt{\sum_{a'}^{} N_{s_t}(a')}} / (1 + N_{s_t}(a))$, and $Q_{s_t}(a) = {W_{s_t}(a)} - L_{s_t}(a)$ with an exploration hyperparameter $c$. The virtual loss is also updated $L_{s_t}(A_t) \gets L_{s_t}(A_t) + \nu$ using a virtual loss hyperparameter $\nu$. The virtual loss prevents the tree search from visiting the same state more than once and discourages the asynchronous workers from all following the same path \cite{segal_2011}.


Once a leaf node, $s_\tau$, is reached, the state is expanded by adding the children of $s_\tau$, $\{ A(s_\tau, a), \forall a \in \mathcal{A}\}$, to the tree $T$. Then, for each child $s'$, the memories are initialized: $W_{s'}(\cdot) = 0$, $N_{s'}(\cdot) = 0$, $P_{s'}(\cdot) = \boldsymbol{p}_{s'}$, and $L_{s'}(\cdot) = 0$, where $\boldsymbol{p}_{s'}$ is the policy output of the network $f_\theta(s')$. Next, the value and policy are computed: $(v_{s_\tau}, \boldsymbol{p}_{s_\tau}) = f_\theta(s_\tau)$ and the value is backed up on all visited states in the simulated path. For $0 \leq t \leq \tau$, the memories are updated: $W_{s_t}(A_t) \gets \max{} (W_{s_t}(A_t), v_{s_\tau}), N_{s_t}(A_t) \gets N_{s_t}(A_t) + 1,  L_{s_t}(A_t) \gets L_{s_t}(A_t) - \nu$. Note that, unlike other implementations of MCTS, only the maximal value encountered along the tree is stored, and not the total value. This is because the Rubik's Cube is deterministic and not adversarial, so we do not need to average our reward when deciding a move.

The simulation is performed until either $s_\tau$ is the solved state or the simulation exceeds a fixed maximum computation time. If $s_\tau$ is the solved state, then the tree $T$ of the simulation is extracted and converted into an undirected graph with unit weights. A full breath-first search is then applied on $T$ to find the shortest predicted path from the starting state to solution. Alternatively, the last sequence $Path = \left\lbrace A_t | 0 \leq t \leq \tau \right\rbrace$ may be used, but this produces longer solutions.

\section{Results}
\begin{wrapfigure}[17]{R}{0.3\textwidth}
    \centering 
    \includegraphics[width=0.29\textwidth]{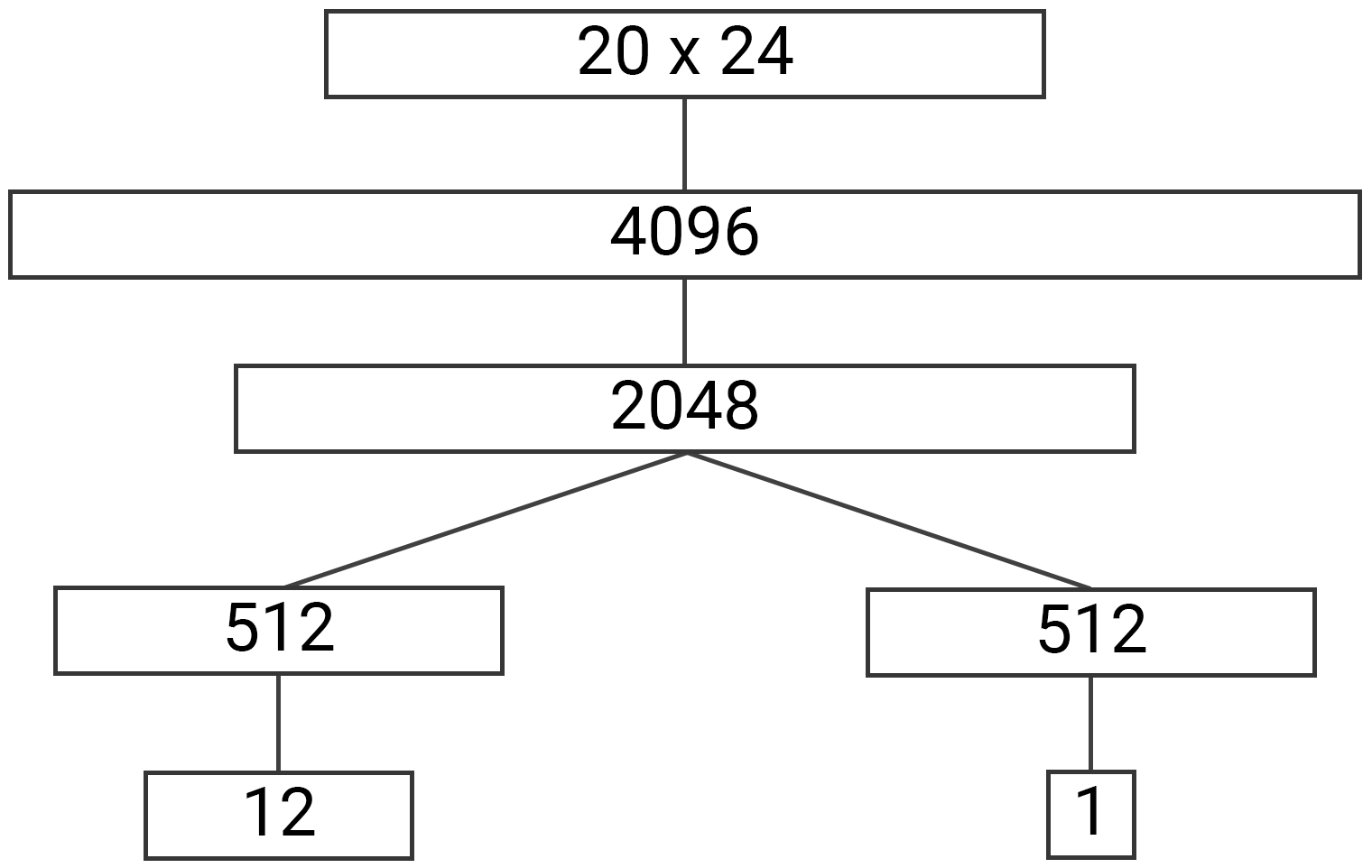}
    \caption{Architecture for $f_\theta$. Each layer is fully connected. We use \textit{elu} activation on all layers except for the outputs. A combined value and policy network results in more efficient training compared to separate networks. \cite{silver_2017}.}
    \label{fig:network}
\end{wrapfigure}

We used a feed forward network as the architecture for $f_\theta$ as shown in Figure \ref{fig:network}. The outputs of the network are a 1 dimensional scalar $v$, representing the value, and a 12 dimensional vector $\boldsymbol{p}$, representing the probability of selecting each of the possible moves. The network was then trained using \AlgoShort{} for 2,000,000 iterations. The network witnessed approximately 8 billion cubes, including repeats, and it trained for a period of 44 hours. Our training machine was a 32-core Intel Xeon E5-2620 server with three NVIDIA Titan XP GPUs.

\captionsetup[subfigure]{width=0.99\textwidth}
\begin{figure}[t]
	\centering
    \begin{subfigure}{0.49\textwidth}
        \includegraphics[width=\textwidth]{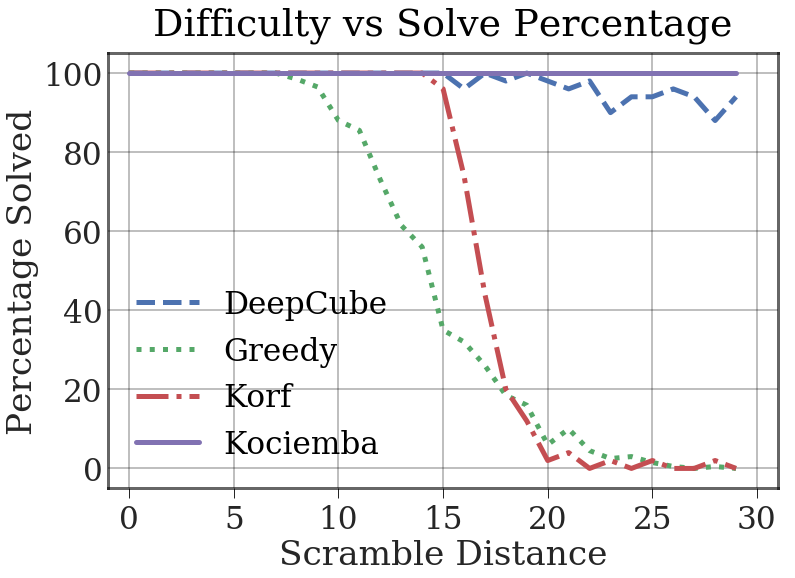}
        \caption{Comparison of solve rates among different solvers. Each solver was given 30 minutes to compute a solution for 50 cubes.}
        \label{fig:perf_time}
    \end{subfigure}
    \begin{subfigure}{0.49\textwidth}
        \includegraphics[width=\textwidth]{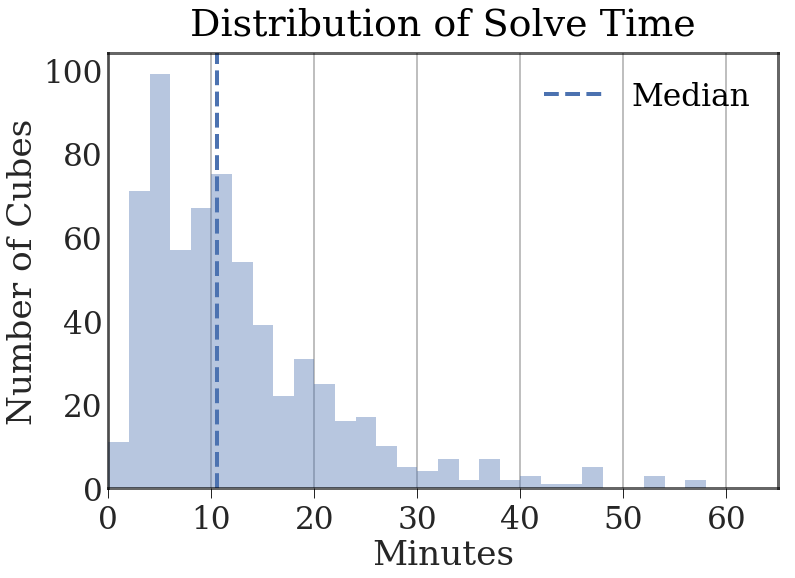}
        \caption{\ProjectName{}'s Distribution of solve times for the 640 fully scrambled cubes. All cubes were solved within the 60 minute limit.}
        \label{fig:solve_time}
    \end{subfigure}
	
    \caption{}
	\label{fig:performance_time}
\end{figure}

As a baseline, we compare \ProjectName{} against two other solvers. The first baseline is the \textbf{Kociemba} two-stage solver \cite{kociemba_2018, tsoy_2018}. This algorithm relies on human domain knowledge of the group theory of the Rubik's Cube. Kociemba will always solve any cube given to it, and it runs very quickly. However, because of its general-purpose nature, it often finds a longer solution compared to the other solvers. The other baseline is the \textbf{Korf} Iterative Deepening A* (IDA*) with a pattern database heuristic \cite{korf_1997,brown_2017}. Korf's algorithm will always find the optimal solution from any given starting state; but, since it is a heuristic tree search, it will often have to explore many different states and it will take a long time to compute a solution. We also compare the full \ProjectName{} solver against two variants of itself. First, we do not calculate the shortest path of our search tree and instead extract the initial path from the MCTS: this will be named \textbf{Naive \ProjectName{}}. We also use our trained value network as a heuristic in a greedy best-first search for a simple evaluation of the value network: this will be named \textbf{Greedy}. An overview of each of the solver's capacity to solve cubes is presented in Figure \ref{fig:perf_time}.

\begin{figure}[t]
	\centering
	\includegraphics[width=0.99\textwidth]{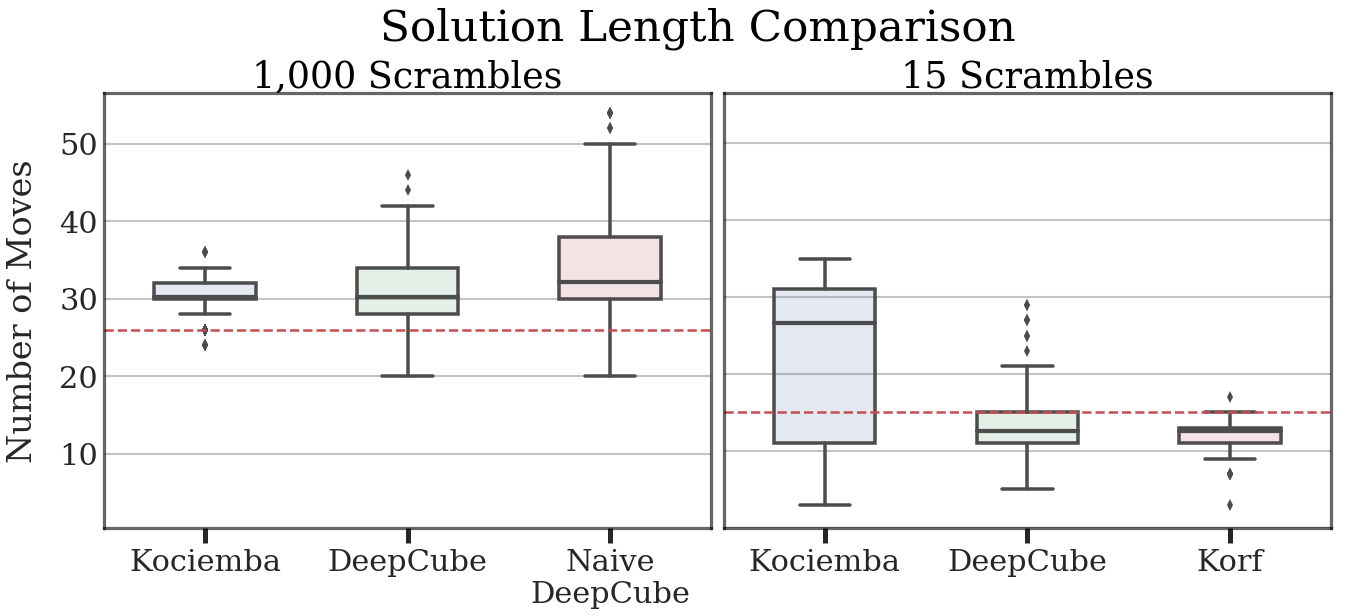}
    \caption{Distribution of solution lengths for \ProjectName{}, Kociemba, and Korf. The left graph features naive \ProjectName{} to evaluate the effect of our shortest path search. The right graph features the Korf optimal solver to evaluate how well \ProjectName{} can find short solutions. The red lines represent the 26 and 15 move upper bound on the left and right respectively.}
	\label{fig:benchmark}
\end{figure}

We compare our results to Kociemba using $640$ randomly scrambled cubes. Starting from the solved cube, each cube was randomly scrambled $1000$ times. Both \ProjectName{} and Kociemba solved all $640$ cubes within one hour. Kociemba solved each cube in under a second, while \ProjectName{} had a median solve time of $10$ minutes. The systematic approach of Kociemba explains its low spread of solution lengths with an interquartile range of only $3$. Although \ProjectName{} has a much higher variance in solution length, it was able to match or beat Kociemba in $55\%$ of cases. We also compare \ProjectName{} against Naive \ProjectName{} to determine the benefit of performing the BFS on our MCTS tree. We find that the BFS has a slight, but consistent, performance gain over the MCTS path ($-3$ median solution length). This is primarily because the BFS is able to remove all cycles from the solution path. A comparison of solution length distributions for these three solvers is presented in the left graph of Figure \ref{fig:benchmark}.

We could not include Korf in the previous comparison because its runtime is prohibitively slow: solving just one of the $640$ cubes took over $6$ days. We instead evaluate the optimality of solutions found by \ProjectName{} by comparing it to Korf on cubes closer to solution. We generated a new set of $100$ cubes that were only scrambled $15$ times. At this distance, all solvers could reliably solve all $100$ cubes within an hour. We compare the length of the solutions found by the different solvers in the right graph of Figure \ref{fig:benchmark}. Noticeably, \ProjectName{} performs much more consistently for close cubes compared to Kociemba, and it almost matches the performance of Korf. The median solve length for both \ProjectName{} and Korf is $13$ moves, and \ProjectName{} matches the optimal path found by Korf in $74\%$ of cases. However, \ProjectName{} seems to have trouble with a small selection of the cubes that results in several solutions being longer than 15 moves. Note that Korf has one outlier that is above 15 moves. This is because Korf is based on the half-turn metric while we are using the quarter-turn metric.

Furthermore, our network also explores far fewer tree nodes when compared to heuristic-based searches. The Korf optimal solver requires an average expansion of 122 billion different nodes for fully scrambled cubes before finding a solution \cite{korf_1997}. Our MCTS algorithm expands an average of only 1,136 nodes with a maximum of 3,425 expanded nodes on the longest running sample. This is why \ProjectName{} is able to solve fully scrambled cubes much faster than Korf.




\subsection{Knowledge Learned}
\begin{wrapfigure}[18]{r}{0.3\textwidth}
    \centering 
    \includegraphics[width=0.28\textwidth]{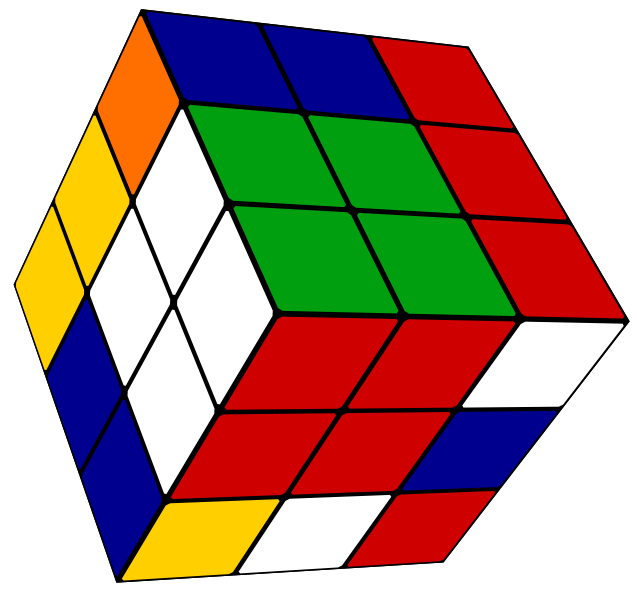}
    \caption{An example of \ProjectName{}'s strategy. On move 17 of 30, \ProjectName{} has created the 2x2x2 corner while grouping adjacent edges and corners together.}
    \label{fig:strategy}
\end{wrapfigure}
\ProjectName{} discovered a notable amount of Rubik's Cube knowledge during its training process, including the knowledge of how to use complex permutation groups and strategies similar to the best human "speed-cubers". For example, \ProjectName{} heavily uses one particular pattern that commonly appears when examining normal subgroups of the cube: $aba^{-1}$. That is, the sequences of moves that perform some action $a$, performs a different action $b$, and then reverses the first action with $a^{-1}$. An intelligent agent should use these conjugations often because it is necessary for manipulating specific cubelets while not affecting the position of other cubelets. 

\begin{figure}[t]
	\centering
	\includegraphics[width=0.99\textwidth]{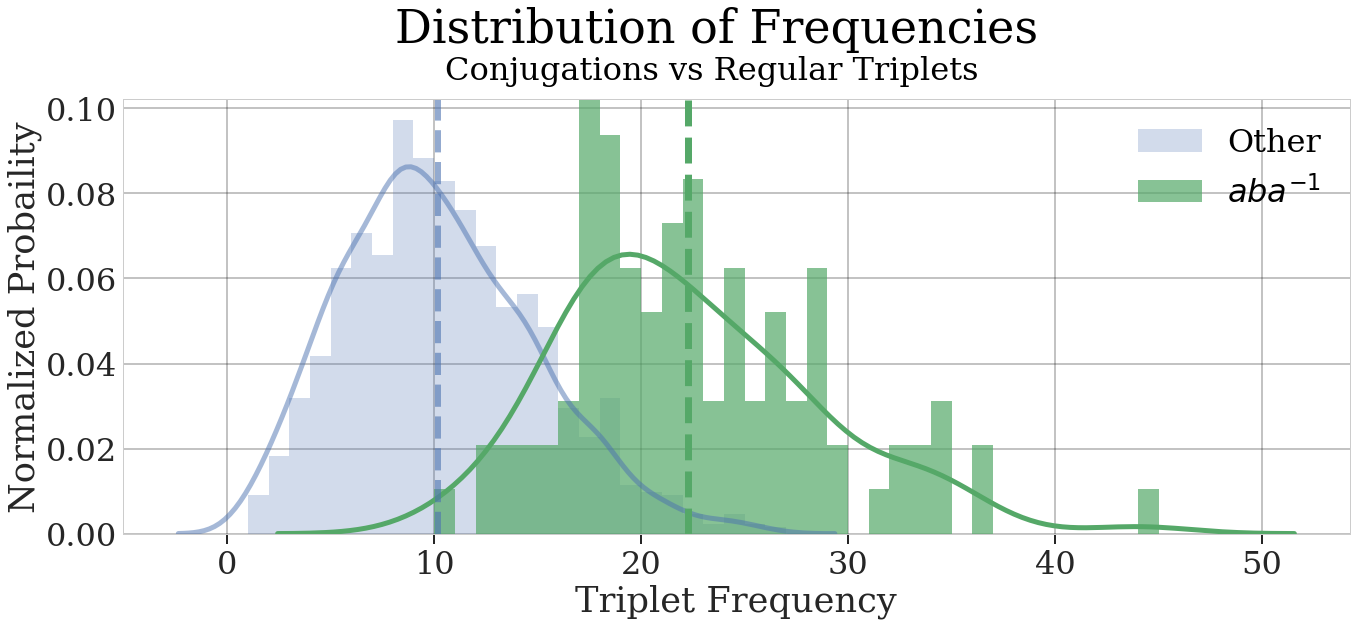}
    \caption{Comparison of the distribution of frequency of the two types of triplets. We split the triplets into conjugations ($aba^{-1}$) and non-conjugations. We then calculate the frequency of each triplet and plot the two distributions. The two vertical lines are the means of their respective distributions.}
	\label{fig:distribution}
\end{figure}

We examine all of the solutions paths that \ProjectName{} generated for the 640 fully scrambled cubes by moving a sliding window across the solutions strings to gather all triplets. We then compute the frequency of each triplet and separate them into two categories: matching the conjugation pattern $aba^{-1}$ and not matching it. We find that the top 14 most used triplets were, in fact, the $aba^{-1}$ conjugation. We also compare the distribution of frequencies for the two types of triplets. In Figure \ref{fig:distribution}, we plot the distribution of frequencies for each of the categories. We notice that conjugations appear consistently more often than the other types of triplets. 

We also examine the strategies that \ProjectName{} learned. Often, the solver first prioritizes completing a 2x2x2 corner of the cube. This will occur approximately at the half way point in the solution. Then, it uses these conjugations to match adjacent edge and corner cubelets in the correct orientation, and it returns to either the same 2x2x2 corner or to an adjacent one. Once each pair of corner-edge pieces is complete, the solver then places them into their final positions and completes the cube. An example of this strategy is presented in Figure \ref{fig:strategy}. This mirrors a strategy that advanced human "speed-cubers" employ when solving the cube, where they prioritize matching together corner and edge cubelets before placing them in their correct locations.

\section{Discussion}


We are currently further improving \ProjectName{} by extending it to harder cubes. 
\AlgoName{} can be used to train a network to solve a 4x4x4 cube and other puzzles such as n-dimensional sequential move puzzles and combination puzzles involving other polyhedra. 

Besides further work with the Rubik’s Cube, we are working on extending this method to find approximate solutions to other combinatorial optimization problems such as prediction of protein tertiary structure. Many combinatorial optimization problems can be thought of as sequential decision making problems, in which case we can use reinforcement learning. Bello et. al. train an RNN through policy gradients to solve simple traveling salesman and knapsack problems \cite{bello}. We believe that harnessing search will lead to better reinforcement learning approaches for combinatorial optimization. For example, in protein folding, we can think of sequentially placing each amino acid in a 3D lattice at each timestep. If we have a model of the environment, \AlgoShort{} can be used to train a value function which looks at a partially completed state and predicts the future reward when finished. This value function can then be combined with MCTS to find approximately optimal conformations.  

Léon Bottou defines reasoning as "algebraically manipulating previously acquired knowledge in order to answer a new question"\cite{bottou_2011}. Many machine learning algorithms do not reason about problems but instead use pattern recognition to perform tasks that are intuitive to humans, such as object recognition. By combining neural networks with symbolic AI, we are able to create algorithms which are able to distill complex environments into knowledge and then reason about that knowledge to solve a problem. \ProjectName{} is able to teach itself how to reason in order to solve a complex environment with only one reward state using pure reinforcement learning. 

In summary, combining MCTS with neural networks is a powerful technique that helps to bridge the gap between symbolic AI and connectionism. Furthermore, it has the potential to provide approximate solutions to a broad class of combinatorial optimization problems. 

\subsubsection*{Acknowledgments}
We thank Harm van Seijen and Yutian Chen for helpful discussions.

\bibliography{mybib}{}
\bibliographystyle{plain}

\end{document}